\documentclass{article}
\usepackage{geometry}
\usepackage[utf8]{inputenc}
\usepackage[english]{babel}
\usepackage{amsmath, amsthm, amssymb}
\usepackage{bbm} % for the doublestroke \mathbbm{1}, indicator functions
\usepackage{xcolor}
\usepackage{authblk}

\usepackage[numbers]{natbib}
\usepackage{graphicx}
\usepackage{makeidx}
\usepackage{todonotes}
\usepackage{url}

\newtheorem{proposition}{Proposition}

\numberwithin{equation}{section}
\theoremstyle{plain}
\usepackage{localMacros}
\makeindex
\begin{document}

\title{Advestis Working Paper \\
Online inference for multiple changepoints and risk assessment}
%%%%%
\author[1]{Olivier Sorba}
\affil[1]{RandomPulse, 75116 Paris, France\\
        olivier.sorba@randompulse.net}
%%%%
\author[2]{C. Geissler}
\affil[2]{Advestis, 69 Boulevard Haussmann, 75008 Paris, France, cgeissler@advestis.com}
%%%%%%%%

\date{April 2021}
%%%%%%%%%%%%%%%%%%%%%
\maketitle
\thanks[1,2]{Work supported by Advestis}
\begin{abstract}
The aim of the present study is to detect abrupt trend changes in the mean of a multidimensional sequential signal. Directly inspired by papers of Fernhead and Liu (\cite{Fearnhead2006} and \cite{FearnheadLiu2007}), this work describes the signal in a hierarchical manner : the change dates of a time segmentation process trigger the renewal of a piece-wise constant emission law. Bayesian posterior information on the change dates and emission parameters is obtained. These estimations can be revised online, i.e. as new data arrive. This paper proposes explicit formulations corresponding to various emission laws, as well as a generalization to the case where only partially observed data are available.
Practical applications include the returns of partially observed multi-asset investment strategies, when only scant prior knowledge of the movers of the returns is at hand, limited to some statistical assumptions. This situation is different from the study of trend changes in the returns of individual assets, where fundamental exogenous information (news, earnings announcements, controversies, etc.) can be used.
\end{abstract}


\section{Introduction}

This document outlines a method for the risk assessment of investment rules with specific attention to regime changes over time.
It is based on the inference method proposed by P.Fearnhead et al. in \cite{Fearnhead2006} and \cite{FearnheadLiu2007}.
The intention of the authors is to have a robust method for detecting changes in a vector emission law assumed to be piecewise stationary, such as the daily performance of a set of assets. The analytical framework proposed by Fernhead et al. (\cite{Fearnhead2006} and \cite{FearnheadLiu2007}) proves to be particularly well suited as it allows for online detection of changes in the vector emission law of asset returns under parsimonious assumptions on the parameters of the emission law. The prior knowledge is limited to the instantaneous probability of occurrence of a change, and to the parameters of a Gaussian distribution the emission law trend is drawn from. The purpose here is not to establish a causal or correlation link between observable variables from the 'real' world, and assets returns. Instead, the goal is to detect with a limited lag, the most likely distribution of changes in the underlying distribution. \\
We slightly generalize the framework of the original paper to the case where only some random components of the performance vector can be observed. In the financial domain, an intended application in finance is to get an up-to-date estimate of an 'intermittent' investment strategy. Such a strategy is governed by decision rules that can be active or inactive at each time and for each invested stock. Estimating the law of returns for such strategies has therefore an additional layer of complexity, compared that of individual stocks. The introduction of random activation moves the model away from a pure econometric model. By this latter term, we mean a set of assumptions explicitly connecting companies' fundamental factors (such as size, profitability, industrial sector, etc.) and their returns. When such assumptions cannot anymore realistically be formulated, the Bayesian approach is very fruitful at providing online estimates for changes in the emission law.

% \begin{figure}[h!]
% \centering
% \includegraphics[scale=0.8]{universe}
% \caption{The Universe}
% \label{fig:universe}
% \end{figure}

\subsection{Learning setup, definitions and notations}
One observes a real possibly multi-dimensional random variable $y_t$ at discrete dates 
$t\in \brackets{1,n}$,
for instance a series of asset performances.
Let $y_{i:j}$ denote the observations from time $i$ to time $j$ included.
We describe the process $y_{1:n}$ by the following multiple changepoint model, based on the assumption that the data before and after a changepoint are independent:
\begin{itemize}
    \item There is an unknown random segmentation (a partition in contiguous segments) 
        ${\mathcal{S}}$ of $\brackets{1,n}$
        so that 
        $\brackets{1,n} =\dot\bigcup_{\tau\in{\mathcal{S}}}\tau$. 
    \item For any two integers $i<j$, we write $i\connectedto j$ for the event 
        $\braces{\exists \tau \in \mathcal{S}, \left[i,j\right) \subset \tau}$, in other words $i, i+1, \cdots, j-1$ belong to the same segment.
    \item To each segment 
        $\tau\in{\mathcal{S}}$
        are associated a tuple $\beta_{\tau}$ of random parameters describing the emission law of 
        $y_{\tau}:= (y_t)_{t\in\tau}$,
        for instance $\beta_{\tau}=\paren{\mu_{\tau}, v_{\tau}}$ where a $\mu_\tau$ is a location parameter and  $v_\tau$ is a size or dispersion parameter.
        We denote 
        $\pi(\beta)\mathrm{d}\beta$,
        $\pi(\mu)\mathrm{d}\mu$,
        $\pi(v)\mathrm{d}v$
        and so on the corresponding prior densities.
    \item Hierarchical structure: 
        \begin{itemize}
            \item conditional on the segmentation ${\mathcal{S}}$, the parameters sets and the observations of different segments are mutually independent, 
            \item the parameter sets follow the same distribution $\pi(\beta)\mathrm{d}\beta$,
            \item the observations follow in each segment $\tau=\left[t,s\right)$ the distribution $\condProbability{y_{t:s-1}}{t\connectedto s,\beta_{\tau}}\mathrm{d}y_t \cdots \mathrm{d}y_{s-1}$.
        \end{itemize}
    \item The starts of the segments of ${\mathcal{S}}$ (except $t=1$) arise from a homogeneous point process on $\mathbb{Z}$ observed on the interval $\brackets{2,n}$.
          This point process is defined by the probability mass function $g(t)$ of the distance between two consecutive changepoints.
\end{itemize}

\paragraph{Objective of this paper}
For any time $t\in \brackets{1,n}$, denote $\beta_t$ the unknown random parameter associated with the unique segment containing $t$.
One particular point of interest is the last one $\beta_n$, and we would like to estimate some characteristics of its distribution conditional to the observations $y_{1:n}$.
We start by restating the method of Fearnhead and Liu in \cite{FearnheadLiu2007}, with some slight changes in notation and conventions.
In particular, in our convention a changepoint is the leftmost point of a segment.

\subsubsection{Prior probability distribution of the segmentations}
Denote $G(\cdot)$ the distribution function of the distance between two successive changepoints. 
then
\begin{align*}
G(t)&=\sum_{s=1}^t g(s)
\end{align*}
Denote $g_0(\cdot)$ the mass distribution function of the distance 
$d=\tau_1-1$ the first changepoint $\tau_1$  after $t=1$ to the origin, which is the residual time in the language of renewal processes.
We know classically that the mass distribution function $g_0(\cdot)$ of this distance has the following expression, quantifying the survival bias:
$$
g_0(d)
=
\frac{\sum_{s=d}^\infty{s}^{-1} s g(s)}{\sum_{s=1}^\infty s g(s)}
=
\frac{1-G(d-1)}{\sum_{s=1}^\infty 1- G(s-1)}.
$$ 
The probability of the segmentation ${\mathcal{S}}=\braces{\brackets{1,\tau_1-1},\brackets{\tau_1,\tau2-1}, \cdots, \brackets{\tau_m,n}}$ is expressed as:
$$g_0(\tau_1-1) \prod_{i=1}^{m-1} g(\tau_{i+1}-\tau_{i}) \paren{1-G(n-\tau_m)}.$$

The authors of \cite{FearnheadLiu2007} suggest a negative binomial distribution. 
With parameters $p$ and $n$, then this is the distribution of the number of independent Bernoulli trials before reaching $n$ successes.
Then classically:
% \todo{\cite{FearnheadLiu2007}
% seems to have a typo, $\binom{t-k}{k-1}$ instead of $\binom{t-1}{k-1}$.  
% }
\begin{align*}
    g(t)
    &=
    \binom{t-1}{n-1} p^n (1-p)^{t-n},\\
    g_{0}(t)
    &=
    \sum_{i=1}^{n}\binom{t-1}{i-1} p^i (1-p)^{t-i}/n.
\end{align*}
\begin{proof}
Consider a Markov process of on $\mathbb{Z}_n$ with $x\rightarrow x$ with probability $1-p$ and $x\rightarrow x+1$ with probability $p$.    Then $g()$ is the probability distribution of the return time to $x=0$ (via $x=n-1$) starting from $0$, and
$g(t)
=p \probability{x_{t-1}=n-1}
=p \probability{\mathcal{B}(t-1,p)=n-1}$.
Additionally, $g_0()$ is the same starting in the process' stationary distribution, which is uniform by cyclicity.
\end{proof} 
For $n=1$ the survival law is geometric and the point process is Markov. Higher values of $n$ can reduce the number of  very short segments.

%%%%%%%%%%%%%%%%%%%%%%%%%%%%%%%%%%%%%%%%%%%%%%%%%%%%%%%%%%%%%%%%
\section{Filtering Recursions}
Let us first define quantities that will appear frequently in subsequent calculations.
\subsection{Segment likelihood}
First define for $t\leq s$ the \textit{ex ante} distribution on segment observations:
\begin{align}\label{eq:defP}
P(t,s)
&=\condProbability{y_{t:s}}{t\connectedto s+1},\\
&=\int\condProbability{y_{t:s}}{t\connectedto s+1, \beta}\pi(\beta)\mathrm{d}\beta
\end{align}
assuming that $P(t,s)$ can be calculated analytically or numerically for all 
$\brackets{t,s}\subset\brackets{1,n}$.
As the authors of \cite{FearnheadLiu2007} point, this requires either conjugate priors on $\beta$ or numerical integration. 
We postpone to section \ref{sec:numericalExamples} the description of concrete tractable examples. 

\subsection{Predecessor changepoint process}
Define $C_t$ as the location of the changepoint immediately preceding $t$, or $1$ if there is none. $C_{n+1}$ refers to the last breakpoint. In other words $C_t$ is the start of the segment containing $t-1$.
The only manner for $C_{t+1}$ to differ from $C_{t}$ is for $t$ to be a changepoint. 
It follows from the model described in introduction that $C_t$ is a Markov process with for $i>1$:
\begin{align}
    \label{eq:CtMarkov}
    \condProbability{C_{t+1}=j}{C_t=i}
    &=
    \begin{cases}
    \frac{1-G(t-i)}
         {1-G(t-i-1)}
         & \text{ if } j=i,\\
    \frac{G(t-i)-G(t-i-1)}{1-G(t-i-1)}& \text{ if } j=t,\\
    0 & \text{otherwise}.
    \end{cases}
\end{align}
The case of $i=1$ is obtained by substituting $G_0(\cdot)$ to $G(\cdot)$ in the relation above.

As usual with hidden Markov models and Viterbi algorithms, the process $C_t$ will follow a reverse Markov chain when conditioned to the observations \cite[chap. 3 p. 51]{cappe2005}.
In this perspective 
define for $1\leq j \leq t < n+1$ the conditional probabilities
\begin{align}
    p_t^{(j)}
    &=
    \condProbability{C_{t+1}=j}{y_{1:t}} .
\end{align}

Assume the posterior distributions $p_t^{(\cdot)}$ are known.
This allows to sample the position of the last changepoint from its exact posterior distribution.
Finally the full segmentation can be sampled from its exact posterior distribution by iterating backwards until the origin of the observations is reached.
In the same manner, maximum a posteriori (MAP) estimates  are obtained by recursively taking the most probable previous changepoint.
A single backward pass yields the marginal posterior changepoint probabilities thanks to the recursion relation:
\begin{align}
    \condProbability{i \,\mathrm{changepoint}}{y_{1:n}}
    &=
    p_n^{(i)}+
    \sum_{j=i+1}^n 
        p_{j-1}^{(i)}
        \condProbability{j \,\mathrm{changepoint}}
                        {y_{1:n}}.
\end{align}

\subsection{Forward recursion on the predecessor changepoint $p_{t}^{(\cdot)}$ distributions}\mbox{}
When $y_1, y_2, \cdots, y_t$ is known and $y_{t+1}$ becomes available, the additional information provided by $y_{t+1}$ is contained in the following likelihood ratios, that Fearnhead and Liu propose to use as update weights for the posterior probability $p_{t}^{(\cdot)}$: 
\begin{align}\label{eq:wtj}
    w_{t+1}^{(j)}
    &:=
    \condProbability{y_{t+1}}{C_{t+2}=j,y_{1:t}},\\
    &=
    \condProbability{y_{t+1}}{j \connectedto t+1 ,y_{j:t}},\\
    &=
    \begin{cases}
    \frac{P(j,t+1)}{P(j,t)} & \text{ if } j<t+1,\\
          P(t+1,t+1)          & \text{ if } j=t+1.\\
    \end{cases}
\end{align} 
In order to do so, the following recursion relation is available:
\begin{align}\label{eq:wtj1}
    p_{t}^{(j)}
    &\propto
        \begin{cases}
        w_t^{(j)} \frac{1-G(t-j)}{1-G(t-j-1)} p_{t-1}^{(j)}
            & \text{ if } j<t,
            \\
        w_t^{(t)}
            \sum_{i=1}^{t-1}
            \frac{G(t-i)-G(t-i-1)}{1-G(t-i-1)}
            p_{t-1}^{(i)}
            & \text{ if } j=t.
        \end{cases}
\end{align}
We recall the demonstration for completeness (\cite{FearnheadLiu2007}) in Section \ref{sec:proof:eq:wtj1}.

These weights usually rely on summary statistics that can be incrementally updated, limiting the processing cost.
\subsubsection{Approximate inference}
In Equation \ref{eq:wtj1} above, the set of indices $\brackets{1, t-1}$ may be seen as a particle swarm of candidate changepoints knowing the signal up to date $t$. 
The authors of \cite{FearnheadLiu2007}
show how to speedup calculations by limiting this particle swarm to the most likely changepoints, based on an efficient re-sampling scheme.
As we deal with time series of moderate length, there is no immediate need for optimization and we only restate the recursion method.
%%%%%%%%%%%%%%%%%%%%%%%%%%%%%%%%%%%%%%%%%%%%%%%%%%%%%%%%%
%%%%%%%%%%%%%%%%%%%%%%%%%%%%%%%%%%%%%%%%%%%%%%%%%%%%%%%%%%%%%%%%%%%%%%%%%%%%%%
\section{Risk evaluation}
After recalling the method proposed by the authors of \cite{FearnheadLiu2007},
we would like to apply the same principles to evaluate the posterior probability of an event $\Omega$ related to the parameters of the last segment, for instance $\Omega_\theta = \braces{\mu_n<\theta}$ for some real $\theta$ assuming that we know how to express the corresponding probability within a connected segment: 
\begin{align}\label{eq:POmegaDef}
    P_{\Omega}(i,j)
    &:=
    \condProbability{\Omega,y_{i:j}}{i\connectedto j+1}.
\end{align}

For the frequent case $\Omega=\braces{\mu_j\leq \theta}$, we denote
\begin{align}\label{eq:PThetaDef}
    P_{\theta}(i,j)
    &:=
    \condProbability{\mu_j\leq \theta,y_{i:j}}{i\connectedto j+1}.
\end{align}

\begin{proposition}\label{prop:risk1}
Consider an event $\Omega$ depending only on the last segment's parameter set  (e.g. $\beta_n=\paren{\mu_n, v_n}$,
Then the following relation holds:
\end{proposition}
\begin{align*}
    \condProbability{\Omega}{y_{1:n}}
    &=
    \sum_{j=1}^{n} 
    p_{n}^{(j)}
    \condProbability{\Omega}{y_{j:n},C_{n+1}=j},\\
    &=
    \sum_{j=1}^{n} 
    p_{n}^{(j)}
    \frac{P_{\Omega}(j,n)}{P(j,n)}
\end{align*}
The proof is given in Section \ref{sec:proof:prop:risk1}
%%%%%%%%%%%%%%%%%%%%%%%%%%%%%%%%%%%%%%%%%%%%%%%%%%%%%%%%%%%%%
\section{Examples with known models}\label{sec:numericalExamples}
In this section we provide worked out examples of models suitable to calculate both the segment likelihood function $P(s,s+k-1):=\condProbability{y_{s:s+k}}{s\connectedto s+k}$ and the distribution $\condProbability{\mu}{y_{s:s+k-1},s\connectedto s+k}$
of the segment parameter $\mu$ conditioned to the observation of $y_{s:s+k-1}$.
Following \cite{FearnheadLiu2007} we favor models where the location parameters and the noise parameters are governed by a common scale parameter $\sigma^2$.

As this section deals with a fixed segment $\brackets{s,s+k-1}$ of length $k$, so may assume that $s=1$ by time invariance and also simply write $y$ for $y_{s:s+k-1}$.

\subsection{Gaussian multilinear regression with fixed variance parameter}
Assume $y_t$ is $d$-dimensional for each $1\leq t \leq n$.
To simplify notation we write $y$ as a flat vector of dimension $k d$
\begin{align}
    y:=\bigoplus_{t=s}^{s+k-1} y_t
\end{align}
On a segment of length $k$, consider the linear regression model:
\begin{align}
    \label{eq:model1}
    y
    &=H \mu + \epsilon,\\
    \epsilon &\sim \mathcal{N}_{k d}(0_{k d}, \sigma^2 \Sigma),\\
    \mu &\sim \mathcal{N}_q(0_q, \sigma^2 D)
\end{align}
where 
\begin{itemize}
\item 
$\sigma^2$ is a variance parameter that we assume fixed for the present section,
\item
$H$ is a $k d\times q$ matrix of $q$ regression vectors with $q\leq d$,
\item
$\Sigma$ is a $k d\times k d$ covariance matrix of rank $k d$,
\item 
and 
$D=Diag(\delta_1^2,\cdots,\delta_q^2)$ is a fixed $q \times q$ positive diagonal matrix, representing the \textit{a priori} size ratios  between the explanatory variables terms and the observation noise.
\end{itemize}
Often one assumes that the noise $\epsilon$ has no time autocorrelation and is identically distributed over time, so that the noise covariance matrix writes as a block diagonal matrix:
\begin{align*}
\sigma^2\Sigma
&=
\idm_k \bigotimes \mathrm{Cov}\paren{\epsilon_s}.
\end{align*}
Contrary to \cite{FearnheadLiu2007}, as our observations are multidimensional, we do not assume the observation noise i.i.d across the signal's dimension at fixed date, hence the presence of the $\Sigma$ covariance matrix. 

In the same manner, if the regression vectors have no time evolution, they write as $k$ repetitions of a bloc of size $d$:
\begin{align*}
H_i
&=
\bigoplus_{t=s}^{s+k-1} h_i,
\end{align*}
where for instance $h_i$ can be a principal component of the $d$-dimensional process $y_t$.

In this setup, classically:

\begin{proposition}[Multilinear setup with fixed noise parameter]
\mbox{ }

\label{prop:multilinearSimple}
the following relations hold:
\begin{align}
P(s,s+k-1\vert  \sigma^2)
&=
\paren{2\pi\sigma^2}^{-\frac{d k}{2}}
\abs{\Sigma}^{-\frac{1}{2}}
\paren{\frac{\abs{M}}{\abs{D}}}^{\frac{1}{2}}
\exp\brackets{-\frac{1}{2 \sigma^2}\norm{y}_P^2},\\
\mu \, \vert \, y,\sigma^2
&\sim
\mathcal{N}_q(\hat\mu, \sigma^2 M),\\
H \mu  \, \vert \, y,\sigma^2
&\sim
\mathcal{N}_{k d}(\hat y, \sigma^2 H M H^T),\\
v^T H \mu  \, \vert \, y,\sigma^2
&\sim
\mathcal{N}(v^T\hat y, \sigma^2 v^T H M H^T v).
\end{align}
where $v$ is any $k d$-dimensional real vector and
\begin{align*}
    &M
    =
    \brackets{H^T\Sigma^{-1}H + D^{-1}}^{-1} ,\\
    &P
    =
    \Sigma^{-1}-\Sigma^{-1}H M H^T\Sigma^{-1},\\
    &\norm{y}_P^2
    =
    y^T P y,\\
    &\hat\mu= M H^T\Sigma^{-1} y,\\
    &\hat y = H \hat\mu =  H M H^T\Sigma^{-1} y.
\end{align*}

\end{proposition}
This set of results is close to \cite[3.2 p54]{MarinRobertBayesianCore2007}, although with non i.i.d noise.
The proof is given for completeness in Section \ref{sec:proof:prop:multilinearSimple} 
%%%%%%%%%%%%%%%%%%%%%%%%%%%%%%%%%%%%%%%%%%%%%%%%%%%%%%%%%%%%%%

\subsection{Gaussian multilinear model with Inverse-Gamma prior on variance}
\label{sec:gaussianInverseGammaNoise}
Still following \cite{FearnheadLiu2007}, assume that in the multilinear model of Section \ref{sec:proof:prop:multilinearSimple} above the variance parameter $\sigma^2$ is not fixed but follows an inverse Gamma law of parameters $\nu/2$ and $\gamma/2$ (see \cite{WikipediaInversGamma}):
\begin{align}\label{eq:sigmaPrior}
    \pi(\sigma^2)\text{d}\sigma^2
    &=
    \frac{1}{\Gamma(\frac{\nu}{2})}
    \paren{\frac{\gamma}{2\sigma^2}}^{\frac{\nu}{2}}
    \exp \paren{-\frac{\gamma}{2 \sigma^2}}
    \sigma^{- 2}\text{d}\sigma^2.
\end{align}

Then:
\begin{proposition}[Multilinear model with inverse Gamma prior on variance]
\label{prop:inverseGammaPriorD1}
\mbox{ }

The following relation holds:
\begin{align}\label{eq:PExprSigmaPrior}
    P(s,s+k-1)
    &=
    \paren{\pi}^{-\frac{d k}{2}}
    \abs{\Sigma}^{-\frac{1}{2}}
    \paren{\frac{\abs{M}}{\abs{D}}}^{\frac{1}{2}}
    \frac{\Gamma(\frac{d k+\nu)}{2}}{\Gamma(\frac{\nu}{2})}
    \frac{\paren{\gamma}^{\frac{\nu}{2}}}
         {\paren{\gamma +\norm{y}_P^2}^{\frac{d k+\nu}{2}}}.
\end{align}

Conditioned to $y$, 
the $\sigma^2\vert y$ random parameter
follows an inverse Gamma law of parameter 
$(\frac{\nu+ k d }{2},\frac{\gamma + \norm{y}_P^2}{2})$.
Notably,  
\begin{align*}
    \condExpectation{\sigma^2}{y}
    &=
    \frac{\gamma + \norm{y}_P^2}{\nu+ k d -2}
    \text{ for } \nu+ k d>2,\\
    %%%%%%%%%%%%%%%%%%%%%%%%%%%%%%%%%%
    \mathbb{V}\brackets{\sigma^2 \sachant y}
    &=
    \frac{2}{\nu+ k d -4}\condExpectation{\sigma^2}{y}^2
    \text{ for } \nu+ k d>4.
\end{align*}

For any $v\in \mathbb{R}^q$ The random variable
$v^T \mu\vert y$
is distributed like 
$v^T \hat\mu+\paren{\frac{\gamma v^T M v}{\nu}}^{\frac{1}{2}} \mathcal{T}_{\nu}$
where $\mathcal{T}_{\nu}$ is a Student's $T$ variable with fractional $\nu$ degrees of freedom.
Notably 
\begin{align}
    \condExpectation{v^T \mu}{y}
    &=
    v^T \hat\mu,\\
    &=
    v^T M \Sigma^{-1} y,\\
    %%%%%%%%%%%%%%%%%%%%%%%%%%%%%%%%%%%%%%%%
    \mathbb{V}\brackets{v^T \mu \sachant y}
    &=
    \frac{\gamma }{\nu-2}v^T M v \text{ if } \nu >2.
\end{align}
Last the law of linear transform $A^T (\mu-\hat \mu) $ of higher rank is a multivariate $t$-distribution (see \cite{WikipediaMultivariateT}) of covariance $\frac{\gamma}{\nu-2} A^T M A $, assuming this last matrix is inversible.
\end{proposition}
The classical Equation \ref{eq:PExprSigmaPrior} is given without proof in \cite{FearnheadLiu2007} in the case $\Sigma=\idm_{k d}$.
This results are also similar to \cite[3.2 p.54]{MarinRobertBayesianCore2007}, still with  $\Sigma=\idm_{k d}$.

%%%%%%%%%%%%%%%%%%%%%%%%%%%%%%%%%%%%%%%%%%%%%%%%%%%%%%%%%%%%%%%%%%%%%%%%%%%%%%%%%%%%%%%%%%%%%%%%%%%%%%%
\subsection{Partially observed multilinear model}
In this section we keep the setup of Section \ref{sec:gaussianInverseGammaNoise} above, but assume the observations are only partial.
More precisely, we assume there is a random  projector $P$ such that only $Py$ is observed.
For instance, the case:
\begin{align*}
    \Pi
    &=
    \bigoplus_{t=s}^{s+k-1} \sum_{i=1}^{d}R_{i,t}e_i e_i^T,
\end{align*} 
where $\paren{e_1,\cdots, e_d}$ is a basis of $\mathbb{R}^d$ and $R_i,t$ a random activation rule.
There is no need to specify independence rules between  $\Pi$ and other variables since we are only interested in what happens in the image of $\Pi$, in other words in the components of $y$ where the rules are activated. Then all estimates can be conditioned to $\Pi$.

In this modified setup, the model of Section \ref{sec:gaussianInverseGammaNoise} is maintained but considered latent, 
the observations being modeled by:
\begin{align*}
    \Pi y 
    &=
    \Pi H \mu + \Pi\epsilon,\\
    \Pi\epsilon 
    &\sim
    \mathcal{N}_{\trace{\Pi}}(0, \sigma^2 \Pi\Sigma \Pi^T),\\
    \mu 
    &\sim
    \mathcal{N}_{q}(0, \sigma^2 D),\\
    \frac{1}{\sigma^2}
    &\sim
    \Gamma(\frac{\gamma}{2},\frac{\nu}{2}).
\end{align*}
% \todo{check $$\gamma/2, \nu/2$$ inversion}
This is exactly the setup of Section \ref{sec:gaussianInverseGammaNoise}
when 
$y$, $H$ and $\Sigma$ are replaced by 
$P y$, $P H$ and $P \Sigma P^T$.
So the same considerations lead to:

\begin{proposition}
conditioned to $\Pi y$ and $\Pi$, the following relations hold:
\begin{align}
    P(s,s+k-1\vert \sigma^2)
    &=
    \paren{2\pi\sigma^2}^{-\frac{\trace{\Pi}}{2}}
    \abs{\Sigma_\Pi}_\Pi^{-\frac{1}{2}}
    \paren{\frac{\abs{M_\Pi}}{\abs{D}}}^{\frac{1}{2}}
    \exp\brackets{-\frac{1}{2 \sigma^2}\norm{y}_{P_\Pi}^2},\\
    %%%%%%%%%%%%%%%%%%%
    P(s,s+k-1)
    &=
    \paren{\pi}^{-\frac{\trace{\Pi}}{2}}
    \abs{\Sigma_\Pi}_{\Pi}^{-\frac{1}{2}}
    \paren{\frac{\abs{M_\Pi}}{\abs{D}}}^{\frac{1}{2}}
    \frac{
            \Gamma\paren{\frac{\trace{\Pi}+\nu}{2}}
         }
         {
           \Gamma\paren{\frac{\nu}{2}}
         }
    \frac{\paren{\gamma}^{\frac{\nu}{2}}}
         {\paren{\gamma +\norm{y}_{P_\Pi}^2}^{\frac{\trace{\Pi}+\nu}{2}}}.
\end{align}
as well as the following relations in law:
\begin{align}
    %%%%%%%%%%%%%%%%%%%%%%%%%%%%%%%%%%%%%%%%%%%%%%%%%%%%%%%%%%%%%%%
    \mu \, \vert \, \sigma^2
    &\sim
    \mathcal{N}_q(\hat\mu, \sigma^2 M_\Pi),\\
    %%%%%%%%%%%%%%%%%%%%%%%
    \sigma^2  
    &\sim
    \operatorname{Inv-Gamma}\paren{\frac{\nu+ \trace{\Pi} }{2},\frac{\gamma + \norm{y}_{P_\Pi}^2}{2}},\\
    %%%%%%%%%%%%%%%%%%%%%%%%%%%%%%%%%%%%%%%%%%%%%%%%%%%%%%%%%%%%%%%%
    v^T \paren{\mu-\hat \mu}  
    &\sim
    \paren{\frac{\gamma v^T M v}{\nu}}^{\frac{1}{2}} \operatorname{Student's-t}(\nu).
\end{align}
where $v \in \mathbb{R}^d$ and 
\begin{align*}
    &\Sigma_{\Pi}
    =
    \Pi \Sigma \Pi^T,\\
    %%%%%%%%%%%%%%%%%%
    &M_{\Pi}
    =
    \brackets{H^T\Sigma_{\Pi}^{+} H + D^{-1}}^{-1} ,\\
    %%%%%%%%%%%%%%%%%%
    &P_{\Pi}
    =
    \Sigma_{\Pi}^{+}
    -\Sigma_{\Pi}^{+} H M_{\Pi} H^T\Sigma_{\Pi}^{+},\\
    %%%%%%%%%%%%%%%%%%
    &\norm{y}_{P_{\Pi}}^2
    =
    y^T P_{\Pi} y,\\
    %%%%%%%%%%%%%%%%%%
    &\hat\mu= M_{\Pi} H^T\Sigma_{\Pi}^{+} y,\\
    %%%%%%%%%%%%%%%%%%
     &\hat y= H M_{\Pi} H^T\Sigma_{\Pi}^{+} y.
    %%%%%%%%%%%%%%%%%%
\end{align*}

\end{proposition}
 
To keep all the matrices of same dimension, we introduced the pseudo inverse
$\Sigma_\Pi^{+}$
obtained with a few bloc matrices manipulations as 
$$\Pi \brackets{\paren{\idm_{k d}-\Pi} + \Pi \Sigma \Pi}^{-1}\Pi$$
 and the determinant 
 $\absj{\Sigma_\Pi}_{\Pi}$
 of its restriction to 
 $Im(\Pi)$
 obtained as 
$$\abs{\paren{\idm_{k d}-\Pi} + \Pi \Sigma \Pi}.$$

Note that with a null $P$, which means no observation, we recover the prior distributions of the segment parameters $\mu$ and $\sigma^2$.

\section{Examples}
In this section we detail the calculation of the segment likelihood function is some common cases.
\subsection{Regression by step functions : time invariant noise and covariates}
A simple case occurs when one assumes the signal's conditional expectation is constant over each segment and one wants to explain the observed signal by step functions. 
then $H$ may be expressed as 
\begin{align}
H
&=
\bigoplus\limits_{t=s}^{s+k-1} H_0,
\end{align}
where $H_0$ is a $d \times q$ matrix of $q$ covariate vectors of dimension $d$.
In addition, let us assume that the distribution of the noise $\epsilon$ is time invariant and presents no cross-correlation over different dates.
Then its correlation matrix may be written as a bloc diagonal matrix:
\begin{align}
    \Sigma
    &=
    \bigoplus\limits_{t=s}^{s+k-1} \Sigma_0,
\end{align}
where $\Sigma_0$ is the noise correlation structure at any date.

Last,
we assume the partial information situation is described by the activations $r_{i,t}$, so:
 \begin{align*}
 \Pi
 &=
 \bigoplus_{t=s}^{s+k-1} \Pi_t,\\
 &=
 \sum\limits_{t=s}^{s+k-1}\sum\limits_{i=1}^{d} r_{i,t} e_{i,t} e_{i,t}^T,
 \end{align*}
 with the same bloc structure than the noise correlation.
 
In this case, some bloc matrix algebra leads to:
\begin{align}
\Sigma_{\Pi}^{+}
&=
\bigoplus_{t=s}^{s+k-1}\Pi_t \brackets{\paren{\idm_{d}-\Pi_t} + \Pi_t \Sigma_0 \Pi_t}^{-1}\Pi_t,
\end{align}
and 
\begin{align}
\abs{\Sigma_{\Pi}}
&=
\prod_{t=s}^{s+k-1}\abs{\paren{\idm_{d}-\Pi_t} + \Pi_t \Sigma_0 \Pi_t}.
\end{align}
where $\Sigma_0$ is the noise correlation structure at any date and $\Pi_t$ the projector on the component space observable at date $t$.
Finally, all the relevant quantities can be expressed as functions of running sums of matrices or vectors:
\begin{align}
M_{\Pi}
&=
\brackets{ D^{-1}+\sum_{t=s}^{s+k-1} H_0^T \paren{\Pi_t \Sigma_0 \Pi_t}^{+} H_0 }^{-1},\\
%%%%%%%%%%%%%%%%%%%%%
\hat\mu
&=
M_{\Pi}\sum_{t=s}^{s+k-1} H_0^T \paren{\Pi_t \Sigma_0 \Pi_t}^{+} y_t,\\
%%%%%%%%%%%%%%%%%%%%%
\hat y 
&= 
\bigoplus _{t=s}^{s+k-1} \hat y_0,\\
%%%%%%%%%%%%%%%%%%%%%
\intertext{with}
%%%%%%%%%%%%%%%%%%%%%
\hat  y_0& = H_0 M_{\Pi} \sum_{t=s}^{s+k-1} H_0^T \paren{\Pi_t \Sigma_0 \Pi_t}^{+} y_t
,\\
%%%%%%%%%%%%%%%%%%%%%
y^T P_{\Pi} y
&=
 \sum_{t=s}^{s+k-1} 
   y_t^T \paren{\Pi_t \Sigma_0 \Pi_t}^{+} y_t 
  - \hat \mu^T M^{-1}\hat \mu,\\
  &=
 \sum_{t=s}^{s+k-1} 
   y_t^T \paren{\Pi_t \Sigma_0 \Pi_t}^{+} y_t 
  -\sum_{t=s}^{s+k-1} 
   \hat  y_0^T \paren{\Pi_t \Sigma_0 \Pi_t}^{+}\hat y_0
  - \hat \mu^T D^{-1}\hat \mu.
\end{align}
% \todo{check 5.9 and develop calculation} 
%%%%%%%%%%%%%%%%%%%%%%%%%%%%%%%%%%%%%%%%%%%%%%%%%%%%%%%%
\subsection{Regression by step functions : time invariant covariates and white noise}
In addition to the preceding section, let us assume first hand that the noise
$\epsilon$ is i.i.d so that $\Sigma_0=\idm_d$, and second that the covariates at any date are the natural basis of $\mathbb{R}^d$, so that $q=d$ and $H_0=\idm_d$.  

Finally, some algebraic transformations lead to
\begin{align*}
    \Sigma_{\Pi}
    &=
    \sum_{t=s}^{s+k-1}\sum_{i=1}^{d} e_{i,t} r_{i,t} e_{i,t}^T,\\
    %%%%%%%%%%%%%%%%%%
    \absj{\Sigma_{\Pi}
          \text{ restricted to }
          \Pi\paren{\mathbb{R}^{k d}}
         }
    &=
    1\\
    %%%%%%%%%%%%%%%%%%
    M_{\Pi}
    &=
    \sum_{i=1}^{q}e_i\frac{\delta_i^2}{1+n_i \delta_i^2} e_i^T,\\
    %%%%%%%%%%%%%%%%%%
    \absj{M_{\Pi}}
    &=
    \prod_{i=1}^{q} \frac{\delta_i^2}{1+n_i \delta_i^2} ,\\
    %%%%%%%%%%%%%%%%%%
    P_{\Pi}
    &=
    \sum_{i,t} e_{i,t} r_{i,t} e_{i,t}^T -\sum_{i,t,t'} e_{i,t}r_{i,t}\frac{\delta_i^2}{1+n_i \delta_i^2} r_{i,t'}e_{i,t'}^T,\\
    %%%%%%%%%%%%%%%%%%
    \norm{y}_{P_{\Pi}}^2
    &=
    \sum_{i,t} r_{i,t} y_{i,t}^2
    -\sum_{i} \frac{\delta_i^2}{1+n_i \delta_i^2} \paren{\sum_{t}r_{i,t} y_{i,t}}^2,\\
    &= \sum_{i} n_i \paren{\bar{y_i^2}-\frac{n_i\delta_i^2}{1+n_i \delta_i^2} \bar{y_i}^2},\\
    &= \sum_{i} n_i \paren{\bar{ y_i^2}-\bar{y_i}^2 +\frac{1}{1+n_i \delta_i^2} \bar{y_i}^2},\\
    %%%%%%%%%%%%%%%%%%
    \hat\mu&
    = 
    \sum_{i=1}^d \paren{1-\frac{1}{1+n_i \delta_i^2}}\bar{y_i} e_i,\\
    \hat y_t
    &= 
    \sum_{i=1}^d \paren{1-\frac{1}{1+n_i \delta_i^2}}\bar{y_i}  e_{i}.
    %%%%%%%%%%%%%%%%%%
\end{align*}
where for $1\leq i\leq q$
\begin{align*}
    n_i 
    &=
    \sum\limits_{t=s}^{s+k-1} r_{i,t},\\
    \bar y_i 
    &=
    \frac{1}{n_i}\sum\limits_{t=s}^{s+k-1} r_{i,t} y_{i,t},\\
    \bar {y_i^2 }
    &=
    \frac{1}{n_i}\sum\limits_{t=s}^{s+k-1} r_{i,t} y^2_{i,t}. 
\end{align*}
As expected, both the likelihood and the linear estimates appear as mixtures between their counterparts arising one hand from the prior Bayesian model and on the other hand from the purely linear regression model.
This example also shows how naturally the formulation above deals with missing values or dates in the time series. 
%%%%%%%%%%%%%%%%%%%%%%%%%%%%%%%%%%%%%%%%%%%%%%%%%%%%%%%%%
%%%%%%%%%%%%%%%%%%%%%%%%%%%%%%%%%%%%%%%%%%%%%%%%%%%%%%%%%

%%%%%%%%%%%%%%%%%%%%%%%%%%%%%%%%%%%%%%%%%%%%%%%%%%%%%%%%%
%%%%%%%%%%%%%%%%%%%%%%%%%%%%%%%%%%%%%%%%%%%%%%%%%%%%%%%%%
\section{Proofs}
\label{sec:proofs}
%%%%%%%%%%%%%%%%%%%%%%%%%%%%%%%%%%%%%%%%%%
\subsection{Proof of Equation \ref{eq:wtj1}}
\label{sec:proof:eq:wtj1}
\begin{proof} 
The first step is to recall that the prior "previous changepoint" process $C_t$ is a Markov chain with the transition probabilities of Equation \ref{eq:CtMarkov}.

Then by Bayes relation the following filtering recursions hold:
\begin{align*}
    \condProbability{C_{t+1}=j}{y_{1:t}}
    &\propto
    \condProbability{y_t}{C_{t+1}=j,y_{1:t-1}}
    \condProbability{C_{t+1}=j}{y_{1:t-1}}
\end{align*}
and
\begin{align*}
    \condProbability{C_{t+1}=j}{y_{1:t-1}}
    &=
    \sum_{i=1}^{t}
    \condProbability{C_{t+1}=j}{C_{t}=i}
    \condProbability{C_t=i}{y_{1:t-1}}
\end{align*}
so that 
\begin{align*}
    \condProbability{C_{t+1}=j}{y_{1:t}}
    \propto&
        \condProbability{y_t}{C_{t+1}=j,y_{1:t-1}}
        \\&\sum_{i=1}^{t-1}
        \condProbability{y_t}{C_{t+1}=j,y_{1:t-1}}
        \condProbability{C_{t+1}=j}{C_{t}=i}
        \condProbability{C_{t}=i}{y_{1:t-1}},\\
    %%%%%%%%%%%%%%%%%%%%%%%
    \propto&
        w_t^{(j)}
        \sum_{i=1}^{t-1}
        \condProbability{C_{t+1}=j}{C_{t}=i}
        \condProbability{C_{t}=i}{y_{1:t-1}}.
\end{align*}
In fine,
\begin{align*}
    \condProbability{C_{t+1}=j}{y_{1:t}}
    &\propto
        \begin{cases}
        w_t^{(j)} \frac{1-G(t-i)}{1-G(t-i-1)} \condProbability{C_{t}=j}{y_{1:t-1}}
            & \text{ if } j<t,
            \\
        w_t^{(j)}
            \sum_{i=1}^{t-1}
            \frac{G(t-i)-G(t-i-1)}{1-G(t-i-1)}
            \condProbability{C_{t}=i}{y_{1:t-1}}
            & \text{ if } j=t,
        \end{cases}
\end{align*}
or equivalently, Equation \ref{eq:wtj1}.
\end{proof}

%%%%%%%%%%%%%%%%%%%%%%%%%%%%%%%%%%%%%%%%%%
%%%%%%%%%%%%%%%%%%%%%%%%%%%%%%%%%%%%%%%%%%%%%

\subsection{Proof of proposition \ref{prop:risk1}}
\label{sec:proof:prop:risk1}
\begin{proof}
Again, we proceed by conditioning on the position of the last changepoint.
Then as the event $\Omega$ only depends on the last segment parameters, 
\begin{align*}
    \condProbability{\Omega}{y_{1:n}}
    &=
    \sum_{j=1}^{n}
    \condProbability{ C_{n+1}=j}{y_{1:n}}
    \condProbability{\Omega}{y_{1:n},C_{n+1}=j},\\
    &=
    \sum_{j=1}^{n}
    p_{n}^{(j)}
    \condProbability{\Omega}{y_{j:n},C_{n+1}=j}.\\
    &=
    \sum_{j=1}^{n} 
    p_{n}^{(j)}
    \frac{\condProbability{\Omega,y_{j:n}}{C_{n+1}=j}}
         {\condProbability{y_{j:n}}{C_{n+1}=j}},\\
    &=
    \sum_{j=1}^{n} 
    p_{n}^{(j)}
    \frac{P_{\Omega}(j,n)}{P(j,n)}.
\end{align*}
\end{proof}

\subsection{Proof of Proposition \ref{prop:multilinearSimple}}
\label{sec:proof:prop:multilinearSimple}
\begin{proof}Recall that the segment parameter set $\beta$ is reduced here to the location parameter $\mu$, and that by definition,
\begin{align*}
    &M
    =
    \brackets{H^T\Sigma^{-1}H + D^{-1}}^{-1} ,\\
    &P
    =
    \Sigma^{-1}\paren{\Sigma-H M H^T}\Sigma^{-1},\\
    &\norm{y}_P^2
    =
    y^T P y,\\
    &\hat\mu= M H^T\Sigma^{-1} y,\\
    &\hat y = H \hat\mu =  H M H^T\Sigma^{-1} y.
\end{align*}

The model definition and a few algebraic transformations based on the definitions above lead to: 
% \todo{verify, add similar reference}
\begin{align}
\condProbability{y}{\mu}\pi(\mu)
&=
\paren{2\pi \sigma^2}^{-\frac{d k}{2}}
\abs{\Sigma}^{-\frac{1}{2}}
\exp\brackets{
    -\frac{1}{2\sigma^2}
    \paren{y-H\mu}^T\Sigma^{-1}\paren{y-H\mu}
}\notag\\
&\;\;\;\;
\paren{2\pi \sigma^2}^{-\frac{q }{2}}
\abs{D}^{-\frac{1}{2}}
\exp\brackets{-\frac{1}{2\sigma^2}\mu^T D^{-1} \mu},\\
%%%%%%%%%%%%%%%%%%%%%%%%%%%%%%%%%%%%%%%%
&=
\paren{2\pi \sigma^2}^{-\frac{d k}{2}}
\abs{\Sigma}^{-\frac{1}{2}}
\exp\brackets{
    -\frac{1}{2\sigma^2}\paren{ y^T \Sigma^{-1} y - 2 y^T \Sigma^{-1} H \mu
}}\notag\\
&\;\;\;\;
\paren{2\pi \sigma^2}^{-\frac{q }{2}}
\abs{D}^{-\frac{1}{2}}
\exp\brackets{-\frac{1}{2\sigma^2}\mu^T M^{-1} \mu},\\
%%%%%%%%%%%%%%%%%%%%%%%%%%%%%%%%%%%%%%%%
%%%%%%%%%%%%%%%%%%%%%%%%%%%%%%%%%%%%%%%%
&=
\paren{2\pi \sigma^2}^{-\frac{d k}{2}}
\abs{\Sigma}^{-\frac{1}{2}}
\exp\brackets{
    -\frac{1}{2\sigma^2}
    \paren{ y^T \Sigma^{-1} y 
    -y^T\Sigma^{-1}H M H^T \Sigma^{-1} y
    }}\notag\\
&\;\;\;\;
\paren{2\pi \sigma^2}^{-\frac{q }{2}}
\abs{D}^{-\frac{1}{2}}
\exp\brackets{-\frac{1}{2\sigma^2}
\paren{\mu-M H^T\Sigma^{-1}y}^T 
M^{-1}
\paren{\mu-M H^T\Sigma^{-1}y}},\\
%%%%%%%%%%%%%%%%%%%%%%%%%%%%%%%%%%%%%%%%
&=
\paren{2\pi \sigma^2}^{-\frac{d k}{2}}
\abs{\Sigma}^{-\frac{1}{2}}
\exp\brackets{
    -\frac{1}{2\sigma^2}
    \norm{y}_P^2
}\notag\\ 
&\;\;\;\;
\paren{2\pi \sigma^2}^{-\frac{q }{2}}
\abs{D}^{-\frac{1}{2}}
\exp\brackets{
    -\frac{1}{2\sigma^2}
    \paren{\mu-\hat{\mu}}^T M^{-1} \paren{\mu-\hat{\mu}}
},
\end{align}
% \todo{(Not clear)  indeed;-)}
The first and second results in the proposition follow by the relation $$P(s,s+k)
=
\int \condProbability{y}{\mu}\pi(\mu)\text{d}\mu.$$ 
and by the Bayesian relation
$$\condProbability{\mu}{y}
\propto
\condProbability{y}{\mu}\pi(\mu).$$
\end{proof}
The third relation follows from the relation $\hat{y}=H\hat{\mu}.$
The last relation follows by a simple expectation and variance calculation

\subsection{Proof of Proposition \ref{prop:inverseGammaPriorD1}}
\begin{proof}
Starting from the results of the fixed $\sigma^2$ model leads to:
\begin{align*}
P(s,s+k-1)
&=
    \iint \condProbability{y}{\mu,\sigma^2}
    \pi(\mu)
    \pi(\sigma^2)
    \text{d}\mu
    \text{d}\sigma^2,\\
&=
    \int 
    \frac{1}{\Gamma(\frac{\nu}{2})}
    \paren{\frac{\gamma}{2\sigma^2}}^{\frac{\nu}{2}}
    \exp \paren{-\frac{\gamma}{2 \sigma^2}}
    \sigma^{- 2}
    P(s,s+k-1\vert \sigma^2)
    \text{d}\sigma^2,\\
&=
    \int 
    \frac{1}{\Gamma(\frac{\nu}{2})}
    \paren{\frac{\gamma}{2\sigma^2}}^{\frac{\nu}{2}}
    \exp \paren{-\frac{\gamma}{2 \sigma^2}}
    \paren{2\pi\sigma^2}^{-\frac{d k}{2}}
    \abs{\Sigma}^{-\frac{1}{2}}
    \paren{\frac{\abs{M}}{\abs{D}}}^{\frac{1}{2}}
    \exp\brackets{-\frac{1}{2 \sigma^2}\norm{y}_P^2}
    \sigma^{- 2}
    \text{d}\sigma^2,\\
&=
    \paren{\pi}^{-\frac{d k}{2}}
    \abs{\Sigma}^{-\frac{1}{2}}
    \paren{\frac{\abs{M}}{\abs{D}}}^{\frac{1}{2}}
    \frac{\Gamma(\frac{d k+\nu)}{2}}{\Gamma(\frac{\nu}{2})}
    \frac{\paren{\gamma}^{\frac{\nu}{2}}}
         {\paren{\gamma +\norm{y}_P^2}^{\frac{d k+\nu}{2}}}.
\end{align*}

Next we know by Bayes' relation combined with the equations above that
\begin{align*}
    \condProbability{\sigma^2}{y}
    &\propto
    \condProbability{y}{\sigma^2}\pi(\sigma^2),\\
    &\propto
    \paren{\sigma^2}^{-\frac{\nu+ k d }{2}}
    \exp{\brackets{-\frac{\gamma + \norm{y}_P^2}{2\sigma^2}}}
    \sigma^{- 2}
    \text{d}\sigma^2,\\
\end{align*}
so that 
$\sigma^2\vert y$
follows an inverse Gamma law of parameter 
$(\frac{\nu+ k d }{2},\frac{\gamma + \norm{y}_P^2}{2})$.
Notably, for $\nu+ k d>2$,  
\begin{align*}
    \condExpectation{\sigma^2}{y}
    &=
    \frac{\gamma + \norm{y}_P^2}{2}
    \frac{\Gamma(\frac{\nu+ k d }{2}-1)}{\Gamma(\frac{\nu+ k d }{2})},\\
    &=
    \frac{\gamma + \norm{y}_P^2}{\nu+ k d -2},
\end{align*}
and for $\nu+ k d>4$,
\begin{align*}
    \mathbb{V}\brackets{\sigma^2 \sachant y}
    &=
    \frac{2}{\nu+ k d -4}\condExpectation{\sigma^2}{y}^2.
\end{align*}

Last, we know from Proposition \ref{prop:multilinearSimple} that conditioned to $y$ and $\sigma^2$,
the random scalar
$v^T H \mu$
is distributed like $\mathcal{N}(v^T H \hat{\mu}, \sigma^2 v^T H M H^T v)$
so 
\begin{align*}
    \condProbability{v^T H \mu=v^T H \hat \mu+z}{y}\mathrm{d}z
    &=
    \int\condProbability{v^T \mu=v^T \hat \mu+z}{y,\sigma^2}
    \pi(\sigma^2)\mathrm{d}\sigma^2\mathrm{d}z,\\
    &\propto
    \int\paren{\sigma^2}^{-\frac{1}{2}}
    \exp{\brackets{-\frac{z^2}{2 \sigma^2 v^T H M H^T v}}}
    \pi(\sigma^2)\mathrm{d}\sigma^2\mathrm{d}z,\\
    &\propto
    \int\paren{\sigma^2}^{-\frac{\nu+1}{2}}
    \exp{\brackets{-\frac{1}{2\sigma^2} \paren{\gamma+\frac{z^2}{ v v^T H M H^T v}}}}
    \sigma^{-2}\mathrm{d}\sigma^2\mathrm{d}z,\\
    &\propto
    \paren{1+\frac{z^2}{ \gamma v^T H M H^T v}}^{-\frac{\nu+1}{2}}\mathrm{d}z,\\
    &\propto
    \paren{1+\frac{1}{\nu}\frac{z^2}{ \nu^{-1}\gamma v^T H M H^T v}}^{-\frac{\nu+1}{2}}\mathrm{d}z,\\
\end{align*}
so 
$z=v^T H(\mu-\hat\mu)\vert y$
is distributed like 
$\paren{\frac{\gamma v^T H M H^T v}{\nu}}^{\frac{1}{2}} \mathcal{T}_{\nu}$
where $\mathcal{T}_{\nu}$ is a Student's $T$ variable with fractional $\nu$ degrees of freedom.
Notably 
\begin{align*}
    \condExpectation{z}{y}
    &=0,\\
    \mathbb{V}\brackets{z \sachant y}
    &=\frac{\gamma }{\nu-2}v^T H M H^T v \text{ if } \nu >2.
\end{align*}
By the same arguments, a linear transform $A^T H \mu $ of higher rank,  will produce a multivariate $t$-distribution (see \cite{WikipediaMultivariateT}) of covariance $\frac{\gamma}{\nu-2} A^T H M H^T A $, assuming this last matrix is invertible.
\end{proof}

%%%%%%%%%%%%%%%%%%%%%%%%%%%%%%%%%%%%%%%%%%%%%%%%%%%%%%%%%%%%%%%%%%%%
\section{Forthcoming applications}
The algorithm described in this paper from the original presentation by Fernhead and Liu \cite{FearnheadLiu2007}, uses particle filter based sampling, in order to limit the number of selected candidates in the determination of breakup instants. This maintains a linear complexity in the number of dates, instead of a quadratic complexity that would be a major obstacle to use for time series with several thousand points. However, this random sampling makes the results of the algorithm themselves random, notably the \textit{maximum a posteriori} estimates. The uncertainty observed in the results increases as the maximum number of candidates is decreased. There is therefore a trade-off between computation time and accuracy. \\
A first study will consist in the evaluation of this trade-off, in other words the search for the minimal allowed number of candidates given a required minimum level of accuracy. The acceptable uncertainty on the \textit{maximum a posteriori} localization of the regime changes can be considered as a business constraint, that will in turn influence the computational performance.\\
A second direction of work concerns the effectiveness of the detection method, compared to commonly used trend measurement methods. A usual context of application of this paper is the online estimation of the average performance of a system, financial or energetic for example. An operator monitors an emission law based on the arrival of new data, and updates the evaluation of the last regime change. Depending on a possible change in the emission law, for example a change in sign of the trend, the operator can modify the scaling factor allocated to this system. It is easy to a posteriori simulate the cumulative effect in time of the actions caused by the detection of regime changes. Different detection methods can thus be compared on an objective basis. This subject will be the subject of a future publication in the field of financial investment strategies.

%%%%%%%%%%%%%%%%%%%%%%%%%%%%%%%%%%%%%%%%%%%%%%%%%%%%%%%%%%%%%%%%%%%%
%%%%%%%%%%%%%%%%%%%%%%%%%%%%%%%%%%%%%%%%%%
%%%%%%%%%%%%%%%%%%%%%%%%%%%%%%%%%%%%%%%%%%%%%%
% \appendix
% \section{Appendix: some additional content}

%%%%%%%%%%%%%%%%%%%%%%%%%%%%%%%%%%%%%%%%%%%%%%%%%%%%%%%%%%%%%%%%%%%
\printindex
%%%%%%%%%%%%%%%%%%%%%%%%%%%%%%%%%%%%%%%%%%%%%%%%%%%%%%%%%%%%%%%%%%%
\tableofcontents
%%%%%%%%%%%%%%%%%%%%%%%%%%%%%%%%%%%%%%%%%%%%%%%%%%%%%%%%%%%%%%%%%%%
\bibliographystyle{plain}
\bibliography{references}
%%%%%%%%%%%%%%%%%%%%%%%%%%%%%%%%%%%%%%%%%%%%%%%%%%%%%%%%%%%%%%%%%%%
% \todototoc
% \listoftodos
%%%%%%%%%%%%%%%%%%%%%%%%%%%%%%%%%%%%%%%%%%%%%%%%%%%%%%%%%%%%%%%%%%%

\end{document}